\documentclass{article}

     \usepackage[final]{neurips_2019}

\usepackage{gensymb}
\usepackage[utf8]{inputenc} % allow utf-8 input
\usepackage[T1]{fontenc}    % use 8-bit T1 fonts
\usepackage{hyperref}       % hyperlinks
\usepackage{url}            % simple URL typesetting
\usepackage{booktabs}       % professional-quality tables
\usepackage{amsfonts}       % blackboard math symbols
\usepackage{nicefrac}       % compact symbols for 1/2, etc.
\usepackage{microtype}      % microtypography
\usepackage{graphicx}
\usepackage{algorithm,algorithmic}
\usepackage{amssymb,mathrsfs,epsfig,epstopdf,amsthm}
\usepackage{mathtools}
\usepackage{physics}
\usepackage{wrapfig}
\title{A Deep Multi-Modal Method for Patient Wound Healing Assessment}

\author{Subba Reddy Oota \textsuperscript{1}, Vijay Rowtula \textsuperscript{1}, 
Shahid Mohammed\textsuperscript{1}\\\textbf{Jeffrey Galitz\textsuperscript{1}}, \textbf{Minghsun Liu\textsuperscript{1}}, \textbf{Manish Gupta\textsuperscript{2}}\\
\textsuperscript{1} Woundtech Innovative Healthcare Solutions, 
\textsuperscript{2} Microsoft AI Research, India \\
{\{soota, vrowtula, shmohammed, jgalitz, mliu\}@woundtech.net, gmanish@microsoft.com} \\}

\begin{document}

\maketitle

\begin{abstract}
Hospitalization of patients is one of the major factors for high wound care costs. %in the US. 
Most patients do not acquire a wound which needs immediate hospitalization.
However, due to factors such as delay in treatment, patient's non-compliance or existing co-morbid conditions, an injury can deteriorate and ultimately lead to patient hospitalization.
In this paper, we propose a deep multi-modal method to predict the patient's risk of hospitalization.
Our goal is to predict the risk confidently by collectively using the wound variables and wound images of the patient.
% we have proposed <Use this in the above statement>
Existing works in this domain have mainly focused on healing trajectories based on distinct wound types.
We developed a transfer learning-based wound assessment solution, which can predict both wound variables from wound images and their healing trajectories, which is our primary contribution.
%We also demonstrate the application of this solution to estimate the number of days for treating a patient's wound, based on their compliance with the wound care system.
We argue that the development of a novel model can help in early detection of the complexities in the wound, which might affect the healing process and also reduce the time spent by a clinician to diagnose the wound.
\end{abstract}

\section{Introduction}
\vspace{-0.2cm}
In recent years, the outstanding performance of deep learning in computer vision-based tasks has resulted into its extensive application for medical image analysis.
%~\citet{}.
The applications include detection, diagnosis, and segmentation of pathology in cancer by~\citet{janowczyk2016deep}, retinal by~\citet{gulshan2016development}, brain by~\citet{nie20163d}, and wound images by~\citet{goyal2018dfunet, garcia2018classification, 8740858, elmogy2018tissues}.
However, fewer practical applications were built that can help clinicians to detect wound types and anticipate the wound's healing trajectory.

In current clinical practice, the evaluation of different wound ulcers 
%such as pressure ulcer, diabetic ulcer, surgical wound, trauma wound, and venous ulcer 
consists of wound specific tasks which help in early diagnosis of wounds. 
Keeping track of development and treatment procedure for each particular case depends on wound type. 
Moreover, many of the wound findings are documented via visual assessment by clinicians.
Some of the challenges for the clinicians in terms of cost of treatment include (i) frequent assessments of a patient, (ii) entry of wound attributes in the database, and (iii) applying the right diagnosis.
%and diminish the valuable time that a clinician could use to help other patients in case of revisiting of a patient.
The clinician's assessment of the wound with its surrounding skin depends on various wound variables such as wound ulcer types, location, stage, margin, yellow slough, and granulation.
We are motivated by the fact that deep learning methods have proven to perform well for object classification and recognition tasks with sufficiently large data.
We tried to automatically assess various factors required for clinician's review, thereby providing semi-supervised wound variable classification.

\begin{figure}[t]
    \begin{center}
        \includegraphics[width=\linewidth]{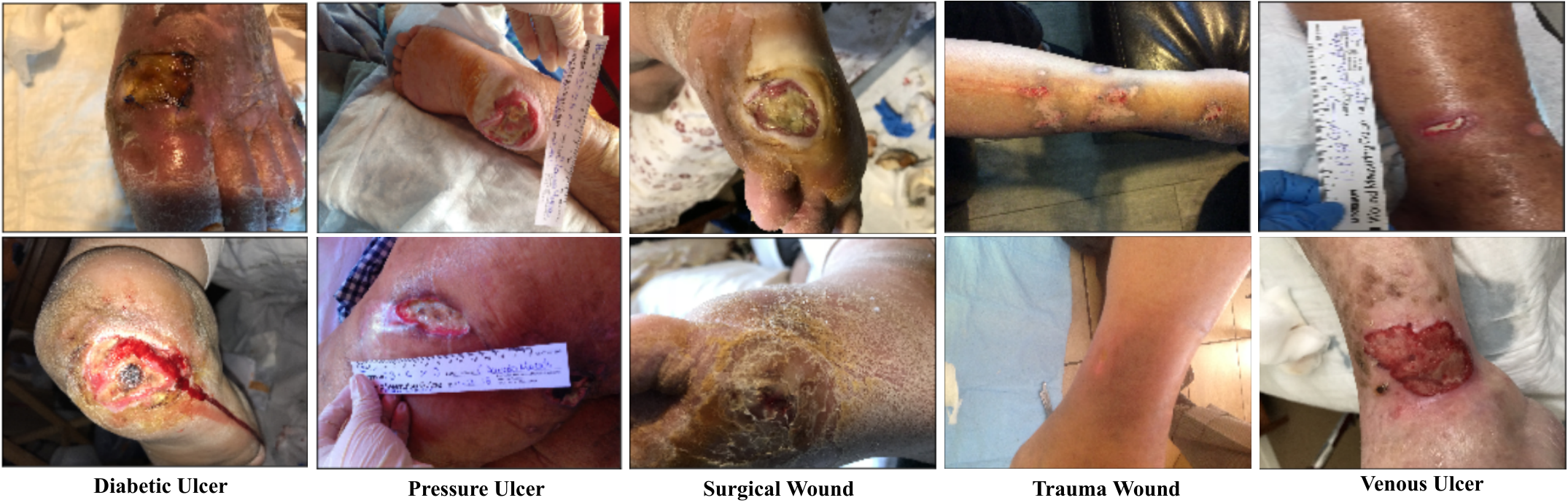}

    \caption{Manually labeled wound ulcer type images}
    %Should be double line. Pls use two images per label, with clear separation between wound types.}
    \label{fig:sampleimages}
    \end{center}
\end{figure}

In this paper, we attempt to detect the wound type and other factors from the wound image. We further investigated the applicability of predicted wound variable to predict patient's hospitalization risk. 
We also show that our models comprising of deep convolutional neural networks (CNNs) provide better performance than a human expert.
Figure~\ref{fig:sampleimages} shows a sample of wound ulcer type images labeled by clinicians.

\section{Our Method}
\vspace{-0.3cm}
The essential aspects of wound assessment comprise of reliable and accurate wound documentation.
%However, due low availability of wound care expertise and higher availability of inexpensive, high-quality image capturing devices along with the current wave of deep learning models for image recognition tasks, it is logical to try and attempt building electronic wound care system.
To provide consistent wound assessments, we implemented a method comprising a mixture of deep neural networks, a simpler machine learning model, and clinicians' expertise to overcome inherent flaws in human wound observations.
Given a wound image as input, the aim is to predict the patient's hospitalization risk. 
To achieve this, we first predict the ``Ulcer Type", ``Location", ``Stage", ``Joint Necrosis", and ``Ligament Necrosis Exposed" attributes of a wound. 
In practice, it is a laborious task to train a convolution neural network model from scratch with randomly initialized weights. 
Moreover, a large number of samples are required to create accurate neural models, as too little data would result in model overfit. 
Here, we employ pre-trained CNN Xception architecture weights~\citep{chollet2017xception} to fine-tune the model on the five heuristic wound variables. 
These five predictions are combined with other 16 clinician-filled variables like ``BMI", ``Tunneling", ``Age", and ``Gender", etc. (by visual observation) to train a multi-modal binary hospitalization risk (or heal/no-heal) LGBM (Light Gradient Boosted Machines) classifier~\citep{ke2017lightgbm}.

%The final step predicts the Risk of Hospitalization by giving as an input ,the features collected from image and clinician, to a LGBM model. 

\subsection{Dataset and Experimental Results}
\paragraph{Wound Dataset:} We experiment with the patients' wound image dataset, which was accumulated from several years of patients' wound care.
To the best of our knowledge, this is the only dataset providing all types of wound variables annotated and verified by clinicians. 
The dataset contains images corresponding to 20 ulcer types with 80\% of the images from 5 types: ``Diabetic Ulcer", ``Pressure Ulcer", ``Surgical Wound", ``Trauma Wound", and ``Venous Ulcer".
The dataset details are displayed in Tables~\ref{ulcertype},~\ref{ulcerlocation}, and~\ref{stage}. 
In this work, we mainly focused on the five wound variables to build the deep learning models.
We used the data provided by clinicians on other wound heuristics to train the final heal/no-heal model jointly with outputs of deep models.
\iffalse
The dataset is split into 80\% of the images for training and 20\% images for testing. 
The other wound variables, including ``Wound Location", ``Wound Stage", ``Wound Margin" are annotated by clinicians as multi-class whereas ``Joint Necrosis Exposed", ``Ligament Necrosis Exposed", ``Muscle Necrosis Exposed" and ``Exudates" are labeled by medical experts with binary masks. 
\fi

{\renewcommand{\arraystretch}{0.5}
\begin{table*}[!htb]
\centering
\footnotesize
\begin{minipage}{.3\linewidth}
\centering
\begin{tabular}{|l|c|} \hline 
\textbf{Ulcer Type} &\textbf{\#Samples}\\ \hline 
 Diabetic Ulcer & 19773  \\ 
 Pressure Ulcer & 47541 \\  
Surgical Wound & 12238 \\ 
Trauma Wound & 13667 \\
 Venous Ulcer & 32492  \\ \hline
\end{tabular}
\caption{Images by Ulcer Type}
\label{ulcertype}
\end{minipage}
\hspace{0.4cm}
\begin{minipage}{.3\linewidth}
\centering
\begin{tabular}{|l|c|} \hline 
\textbf{Wound Location} &\textbf{\#Samples}\\ \hline 
 Lower Leg & 31775  \\ 
 Sacral & 20501  \\  
Foot & 12753   \\ 
Heel & 11226  \\
Ankle & 10375  \\ 
GreatToe & 4547\\\hline 
\end{tabular}
\caption{Images by Location}
\label{ulcerlocation}
\end{minipage}
\hspace{0.6cm}
\begin{minipage}{.3\linewidth}
\centering
\begin{tabular}{|l|c|} \hline 
 \textbf{Ulcer Stage} &\textbf{\#Samples}\\ \hline 
 Full Thickness & 47849  \\ 
 Grade 2 & 20501  \\  
Stage-3  & 12753   \\ 
Stage-4 & 11226  \\
Unstageable & 10375  \\\hline 
\end{tabular}
\caption{Images by Stage}
\label{stage}
\end{minipage}
\end{table*}}

\paragraph{Challenges:}
We faced the following challenges during the training process: (i) occlusion - wound blocked by either scale or doctor's hand when taking the picture, (ii) illumination - wound images have been captured from smartphone in different lighting conditions, (iii) imbalanced data - not all ulcers are having equal number of samples, (iv) similarity - images in one of the wound ulcer looks very similar to images of the other wound ulcers, and (v) deformation - same wound image appears in different forms. 

\iffalse
\paragraph{\noindent{\underline{\textbf{Data Augmentation}}}}
\label{augmentation}
To overcome the limitation of class imbalance problem we applied the augmentation to the wound variables where the variables are having less number of samples. 
We used several image processing techniques like rotation, flipping, and random scaling to perform augmentation.
The rotation was performed on the images at different angles $90\degree$, $180\degree$, and $270\degree$.
We use two types of flipping (horizontal, and vertical) performed on original images.
Since, we used the CNN Xception~\citet{chollet2017xception} architecture to train the model where each image is resized to a size of (299x299).
\fi

%\section{Experimental Setup}
%\label{sec:experiments}

\paragraph{Experimental Setup:} We trained the heal/no heal model in two-steps.
We first created five different CNN models created for the wound variables such as ``Wound Ulcer Type", ``Wound Location", ``Wound Stage", ``Joint Necrosis Exposed", and ``Bone Necrosis Exposed" using our wound image dataset.
We created single-task models on Wound Ulcer Type, and Location, whereas multi-task models on remaining factors (Wound Stage, Joint Necrosis Exposed, and Bone Necrosis Exposed) due to their dependency on Wound Ulcer Type.
The dataset was split into 70\% for training, 10\% for validation, and 20\% for testing. 
We perform a 5-fold cross-validation technique for each wound variable. 
In the second step, we create a heal/no heal model using the five features extracted from the first step and other features like ``Wound Area", ``Wound Volume", ``BMI", ``Patient's Age", etc. filled by a clinician. 
Due to space constraints, we only discuss details of the wound ulcer type prediction model and the heal/no-heal model.

\begin{figure}[t]
    \begin{center}
        \includegraphics[width=0.9\linewidth]{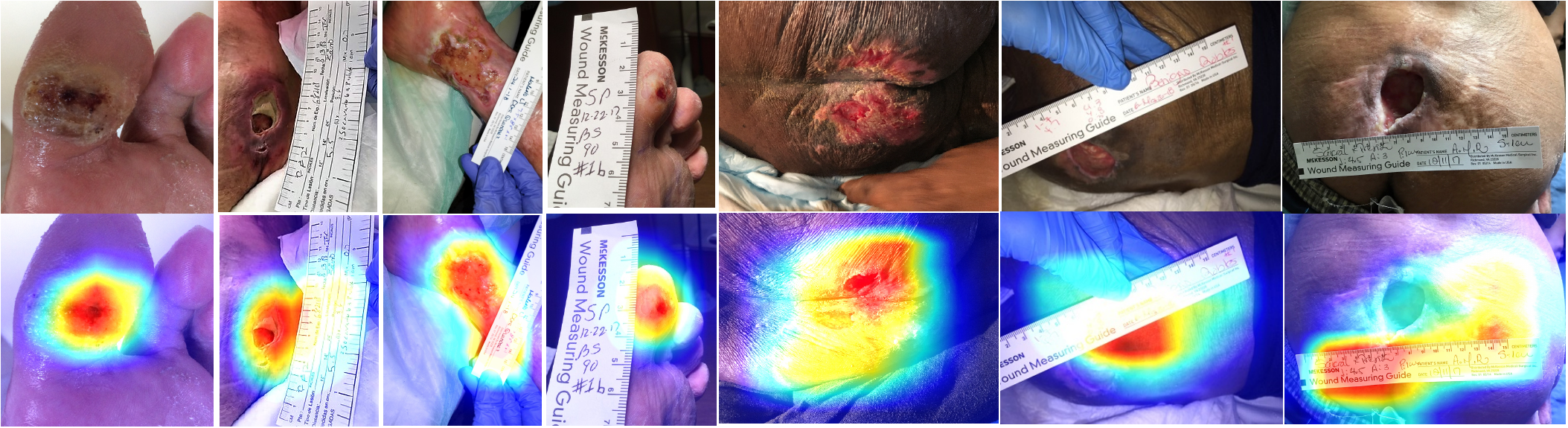}
    \caption{Attention Heatmaps generated by CNN}
    \label{fig:accuracy}
    \end{center}
\end{figure}

%\subsection{Single Task Models}
\noindent\textbf{Wound Ulcer Type Prediction}
To create the wound ulcer type model on wound image dataset, we used a pre-trained Xception architecture and fine-tuned the model. 
We used softmax based output classifier as the prediction layer with a size equal to the number of class labels.
The dataset details are provided in Table~\ref{ulcertype}, where we can observe the class imbalance problem even within the five classes.
We applied the dataset augmentation pre-processing step to deal with the problem. 
For the CNN-Xception model, we set the number of epochs to 50, batch size to 32. We use  Adadelta optimizer~\citet{zeiler2012adadelta} with a learning rate of 0.001.
We illustrate the performance of our model in Table~\ref{ulcertypeaccurcay}.
To interpret the performance of our model, we generated attention heatmaps on wound images using the last CNN layer of Xception, as shown in Figure~\ref{fig:accuracy}.
Although the images used for training have no masking and wound boundary drawn,  Figure~\ref{fig:accuracy} shows that the model can accurately locate the salient information on the wound image.

\iffalse
\paragraph{Wound Location Prediction}
Similar to ulcer type prediction, we created the wound location model on wound image dataset with the pre-trained Xception architecture where a softmax based output classifier used as the last layer with six classes shown in Table~\ref{ulcerlocation}, and sample images shown in Figure~\ref{fig:sampleimages_location}.
Table~\ref{ulcertype} describe the dataset details where the majority of the ulcers formed in these locations.
We drop the samples where the wound occurred on locations which are rare in nature.
Pre-processing and parameters used the same as the ulcer type model and wound location results on the test dataset illustrated in Table~\ref{ulcerlocationaccuracy}.
\fi

{\renewcommand{\arraystretch}{0.5}
\begin{table*}[!htb]
\footnotesize
\centering
\begin{minipage}{.45\linewidth}
\centering
\begin{tabular}{|l|c|c|c|} \hline
\textbf{Ulcer Type} &\textbf{Prec}  &\textbf{Rec}  &\textbf{F1-score}\\ \hline 
 Diabetic Ulcer &0.79 & 0.84 & 0.81 \\ 
 Pressure Ulcer &0.87 & 0.89 & 0.88 \\  
Surgical Wound &0.76 & 0.65 & 0.70 \\ 
Trauma Wound &0.65 & 0.56& 0.61 \\
 Venous Ulcer &0.82 & 0.89 & 0.85  \\ \hline 
\end{tabular}
\caption{Ulcer Type Results}
\label{ulcertypeaccurcay}
\end{minipage}
\hspace{0.5cm}
\begin{minipage}{.45\linewidth}
\centering
\begin{tabular}{|l|c|c|c|} \hline
\textbf{Ulcer Location} &\textbf{Prec} &\textbf{Rec}  &\textbf{F1-score} \\ \hline 
 Lower Leg &0.88 &0.90 & 0.89 \\ 
 Sacral &0.99 & 0.98 & 0.98\\  
Foot &0.83 &0.83 & 0.83  \\ 
Heel &0.84 &0.88 & 0.86 \\
Ankle &0.73 &0.77 & 0.75 \\ 
GreatToe&0.67 &0.55 & 0.61\\\hline
\end{tabular}
\caption{Wound Location Results}
\label{ulcerlocationaccuracy}
\end{minipage}
\end{table*}}

\noindent\textbf{Heal/No-Heal Model}
The main objective of this paper is to create a heal/No-heal model, where we collect five of the wound variables extracted from wound images and other remaining variables filled by the clinician.
To create a heal/no-heal model, we use a recent successful state-of-the-art LightGBM~\citet{ke2017lightgbm} model to classify wound into ``Risk of Hospitalization", or ``Treatment Complete". 
We follow the survival model conditions~\citet{wang2019machine} where we cover the ``Patient Demographic details", ``Procedures", ``Medications", ``Laboratory/Diagnosis of Wound condition'' along with deep model predictions to create the final model. 
Our model provides a precision of 0.68, recall of 0.91, and an F1 of 0.78 for the healing class. 
While for the no-heal class, we obtain a precision of 0.99, recall of 0.79, and an F1 of 0.88.

{\renewcommand{\arraystretch}{0.5}
\begin{table*}[!htb]
\centering
\footnotesize
\begin{minipage}{.45\linewidth}
\centering
\begin{tabular}{|l|c|c|c|} \hline
\textbf{Class} &\textbf{Prec}  &\textbf{Rec}  &\textbf{F1-score}\\ \hline 
 Hospitalization-Wound Related &0.68 & 0.91 & 0.78 \\ 
 Treatment Complete (In active) &0.99 & 0.79 & 0.88  \\  \hline
\end{tabular}
\vspace{-0.2cm}
\caption{Heal/No-Heal Model Results}
\label{healaccurcay}
\end{minipage}
\end{table*}}
%To evaluate our model, we use precision, recall, and F1-score which are summarized in Table~\ref{healaccurcay}.
\vspace{-0.3cm}
\section{Conclusion}
\vspace{-0.2cm}
In this paper, we designed a method to identify the patient hospitalization risk using both wound images and wound attributes provided by clinician. 
Our results show that the method leads to high accuracy for the task.  

\iffalse
\paragraph{Results and Discussion}
To interpret the performance of our model, we generated heatmaps on wound images using last CNN layer of Xception as shown in Figure~\ref{fig:accuracy}.
Although, the images used in training have no masking and wound boundary drawn, the observations from the Figure~\ref{fig:accuracy} that model correctly locate the salient information on the wound image. 
\label{sec:results}

\paragraph{Transfer Learning on NIR (Near Infra-Red images}
To validate our wound ulcer type model, we applied transfer learning on unlabeled dataset of NIR (Near Infra-Red images) .
Here, the labels are validated by clinicians and we are showcasing some sample images in Figure~\ref{fig:ir_nir}.
From the Figure~\ref{fig:ir_nir}, we observe that wound ulcer type model able to predict wound type on near infra-red images for the $1^{st}$, $2^{nd}$, and $4^{th}$ image, whereas wrongly predicted on $3^{rd}$ image as ``Pressure Ulcer" instead of ``Diabetic".

\fi

\bibliographystyle{neurips_2019}
\bibliography{nips_2019}

\end{document}